\documentclass[conference]{IEEEtran}
\IEEEoverridecommandlockouts
\usepackage{cite}
\usepackage{amsmath,amssymb,amsfonts}
\usepackage{algorithmic}
\usepackage{graphicx}
\usepackage{textcomp}
\usepackage{xcolor}
\usepackage{dblfloatfix}
\usepackage{float}
\def\BibTeX{{\rm B\kern-.05em{\sc i\kern-.025em b}\kern-.08em
    T\kern-.1667em\lower.7ex\hbox{E}\kern-.125emX}}

\makeatletter
\newcommand{\linebreakand}{%
  \end{@IEEEauthorhalign}
  \hfill\mbox{}\par
  \mbox{}\hfill\begin{@IEEEauthorhalign}
}
\makeatother

\begin{document}

\title{Predicting Solar Energy Generation with Machine Learning based on AQI and Weather Features\\
}

\author{
\IEEEauthorblockN{Arjun Shah}
\IEEEauthorblockA{\textit{Synapse, Computer Engineering} \\
\textit{D.J. Sanghvi College of Engineering}\\
Mumbai, India\\
arjun.a.shah244@gmail.com}
\and
\IEEEauthorblockN{Varun Viswanath}
\IEEEauthorblockA{\textit{Synapse, Computer Engineering} \\
\textit{D.J. Sanghvi College of Engineering}\\
Mumbai, India\\
varunvis2903@gmail.com}
\and
\IEEEauthorblockN{Kashish Gandhi}
\IEEEauthorblockA{\textit{Synapse, Computer Engineering} \\
\textit{D.J. Sanghvi College of Engineering}\\
Mumbai, India\\
kashishgandhi6112003@gmail.com}
\linebreakand
\IEEEauthorblockN{Dr. Nilesh Madhukar Patil}
\IEEEauthorblockA{\textit{Synapse, Computer Engineering} \\
\textit{D.J. Sanghvi College of Engineering}\\
Mumbai, India\\
nilesh.p@djsce.ac.in}
}

\maketitle 

\begin{abstract}
This paper addresses the pressing need for an accurate solar energy prediction model, which is crucial for efficient grid integration. We explore the influence of the Air Quality Index and weather features on solar energy generation, employing advanced Machine Learning and Deep Learning techniques. Our methodology uses time series modeling and makes novel use of power transform normalization and zero-inflated modeling. Various Machine Learning algorithms and Conv2D Long Short-Term Memory model based Deep Learning models are applied to these transformations for precise predictions. Results underscore the effectiveness of our approach, demonstrating enhanced prediction accuracy with Air Quality Index and weather features. We achieved a 0.9691 R\(^2\) Score, 0.18 MAE, 0.10 RMSE with Conv2D Long Short-Term Memory model, showcasing the power transform technique's innovation in enhancing time series forecasting for solar energy generation. Such results help our research contribute valuable insights to the synergy between Air Quality Index, weather features, and Deep Learning techniques for solar energy prediction. \\

\end{abstract}

\begin{IEEEkeywords}

Solar Power Generation, Zero Inflated Model, Power Transform, Time series, LSTM, Deep Learning
\end{IEEEkeywords}

\section{Introduction}
In the modern world, it has become increasingly clear that eliminating fossil fuels is one of the huge requirements to achieve a carbon-neutral future. The Working Group III Special Report on Renewable Energy Sources and Climate Change Mitigation (SRREN) [19]
suggests that consumption of fossil fuels accounts for the majority of anthropogenic GHG emissions worldwide. It states that CO$_2$ consumption had risen to over 390 ppm, which was around 39\% above pre-industrial levels by the end of 2010. In the race to an efficient energy ecosystem, solar energy is a promising renewable resource, but its intermittent nature poses challenges for integration into the grid. A recent study shows that India lost 29\% of its utilizable global horizontal irradiance potential due to air pollution in the period between 2008 and 2018 [3]. Effective prediction of solar power generation is crucial for efficient planning and management of solar resources. Renewable energy like solar power is said to benefit human beings in a lot of different ways and the most important is in the health domain. Research by Galimova et al. [4] suggests that by 2050 if the world goes under a global transition and the energy sector emissions drop by 92\% then we can reduce premature deaths by air pollution by 97\%. This study reinforced the significance of considering environmental factors in our solar energy forecasting models. This study examines the use of machine learning algorithms that incorporate Air Quality Index (AQI) and meteorological features to improve forecast accuracy.
\\
\\
In order to shed some light on the inconsistent patterns of solar generation data,
a number of regression models were initially utilised to predict the per-hour generation of solar power.
We thus benchmarked  a number of regression models, of which the chief ones were Linear Regression, Lasso, Ridge, ElasticNet, and ensemble models like RandomForest and XGBoost. These models utilize different methodologies which were compared to determine the model with optimal performance to provide valuable insights into the future of solar generation.[1] Additionally, this paper describes a few novel methods such as implementing a Zero Inflated Model and scaling the data using Power Transform, which can significantly improve solar power predictions irrespective of the irregularity in data. This study also incorporates the Convolutional Long Short-Term Memory 2D (ConvLSTM2D) network in the prediction process. ConvLSTM2D is a deep learning algorithm that combines the spatial processing capabilities of Convolutional Neural Networks (CNNs) with the sequential processing capabilities of Long Short-Term Memory (LSTMs)s. Using the spatiotemporal dependence of solar generation data, the ConvLSTM2D network improves the solar energy generation forecast accuracy by building upon historical data. 
\\
\\
The novel contributions of this work can be summarized as follows:
This study hopes to investigate and optimise the performance of the aforementioned machine learning algorithms and provide a clear and accurate picture of solar generation by considering weather features and AQI data which are theorized to have an impact on the fluctuation of this data. It is envisaged that the methodologies employed in this paper will contribute considerably to painting a clearer picture of the sporadic nature of solar power and its influencing factors. The thorough benchmarking and prediction pipeline provides significant benefits for the efficient and sustainable utilization of solar resources by stakeholders, contributing to the adoption and use of practices that ultimately power our society into a clean energy future.

\section{Survey of Literature}

Various studies have used machine learning algorithms to increase the understanding and improvement of solar power forecasting models. Chuluunsaikhan et al. [1] discusses the importance of considering environmental factors such as climate and air pollution when predicting solar power generation. It states that solar panels work best when there is sunlight and no partial shade. However, factors such as weather conditions (e.g. clouds or rain) and air pollution (e.g. fine dust) can cause partial shading and reduce the power output of solar panels. The authors propose a method to regulate the power output of solar panels through machine learning. Machine learning models are developed with three components: weather components, air pollution components, and combined meteorological and air pollution components. The datasets used in the study were collected from 2017 to 2019 from the Seoul province of South Korea. The paper describes the methodology used, including data acquisition, feature extraction, model training, and power output prediction. The authors compare machine learning models, such as linear regression, k-Nearest Neighbors (kNN), Support Vector Regression (SVR), Multi-Level Perceptron (MLP), Random Forest Regressor (RFR), and Gradient-Boosting Regressor (GBR). Models are evaluated using quantitative error methods such as MAE, Coefficient of Determination (R2), and Root Mean Square Error (RMSE). Experimental results show that weather and air pollution parameters can be effective predictors. This paper has been the main premise of our research.
\\
\\
Zhou et al. [2] presents a different approach to forecasting short-term solar power output in smart cities by using deep learning techniques. They used a combination of clustering, CNN, LSTM, and attention mechanisms that obtained improved accuracy in predicting future energy generation. The authors proposed that for future work one could develop training models with time series-based data for further improvement and this is a proposal that we took under consideration. This study and research by Zhou et al. [9] motivated us to incorporate air quality index as a feature in our machine learning models since they [9] used community multiscale air quality in their research indicating how air pollutants can contribute to soiling of PV panels affecting the solar power generation which is also mentioned by Chiteka et al. [12]. Additionally, Jia et al. 

The study in [5] presented models to predict solar radiation; even though our research is based on solar power generation this paper gave us important insights regarding the use of machine learning models in solar forecasting under various weather conditions. Along with this we also considered how machine learning models are computationally better than physical modeling methods since we can use historical data to train the model and predict new data which is difficult to do in the physical modeling approach. In addition, the study by Jebli, Liu, Sweerts et al. [6][7][8] helped us understand the importance of addressing topics like Pearson correlation, air pollutant deposition effects, and random forest optimization in our research. Incorporating this increased the accuracy of the prediction models clearly indicating how different factors and approaches combined can enhance solar power generation prediction. 
\\
\\
Along with machine learning models, there were a lot of studies that suggested the use of deep learning methods for predicting solar power generation. Application of models like CNN's and Recurrent Neural Network (RNN)'s exhibits the effectiveness of these deep learning techniques in capturing complex patterns and dependencies in solar generation signified by Lee et al. [11]. The study by Zazoum et al. [10] also evaluates the accuracy and reliability of deep learning methods in forecasting solar PV power generation which is essential for effective grid integration and energy management. To address the unique characteristics of the dataset, which exhibited an excess of zero values, the researchers in the study proposed by Kim et al. 

Researchers in [13] explored alternative statistical models beyond the traditional Poisson regression. In addition to the Poisson model, the zero-inflated model was employed, acknowledging its ability to effectively handle datasets with an excessive proportion of observed zero values. By employing the zero-inflated model, the researchers sought to capture the dual processes contributing to the occurrence of zeros, distinguishing between structural zeros and excess zeros. Thomas et al.The work in [14] also emphasizes the need for a modeling framework for univariate and multivariate zero-inflated time series of counts. The basic modeling framework used is observation-driven Poisson regression with a Generalized Linear Model (GLM) structure. 
\\
\\
The Zero-Inflated Poisson (ZIP) model is employed to capture the possibility of extra observed zeros relative to the Poisson distribution, a common feature in count data. Using these insights we also utilized zero-inflated models in our research. Yeom et al. [15] introduces a novel approach using deep learning models, specifically ConvLSTM networks, to predict short-term solar radiation by incorporating geostationary satellite images. The proposed model showed high accuracy in capturing cloud-induced variations in ground-level solar radiation compared to the conventional Artificial Neural Network (ANN) and RFR models. This paper led to us implementing ConvLSTM2D on our dataset too. 
\\
\\
To summarize, the reviewed papers have considerably contributed to solar power generation using machine learning and deep learning techniques. Their research provided observations that helped us build our research on and further enhance solar forecasting by utilizing AQI, time series-based data, exploring novel approaches, and other different approaches to making solar power forecasting more reliable and accurate.

\section{Methodology}

\subsection{Dataset}\label{AA}
In order to accurately train our model on features that would help it effectively predict solar generation, we needed a dataset that had high granularity, solar generation,  irradiance, and meteorological data. We thus utilized the UNISOLAR Solar Generation Dataset which includes two years of Photovoltaic solar energy generation data collected at an interval of 15 minutes at La Trobe University Campus in Victoria, Australia. Weather data like apparent temperature, air temperature, dew point temperature, wind speed, wind direction, and relative humidity were also provided by the dataset. We curated the data by merging and selecting provided features such that a suitable dataset may be created for our model. We did this by merging the data provided about the solar panels such as the number of panels and type of inverters, with the aforementioned weather data features to obtain a comprehensive and feature rich data of potential factors affecting solar data. In order to improve the input features of our model, we utilized AQI data sourced from the aqicn.org website, which was captured by a station located in Macleod, Victoria with the location showed in Fig. 1. This station was at a distance of 1.77 km from the University where the UNISOLAR dataset was collected (as shown in Fig. 2).
\\
It was noticed that the solar data generated for 15-minute intervals had a number of irregularities in the number of data collections per day. Switching to one-hour intervals eliminated this problem and made the data intervals regular.
Another reason for the use of 1 hour intervals was to reduce the noise caused due to various external factors like equipment sensitivity, cloud cover and other atmospheric factors.
\\
For our study, we implemented 70:30 dataset split for training and testing respectively. This approach with 70\% of the dataset dedicated to training, allows our model to effectively learn the complex temporal dependencies and relationships inherent in solar data, thereby enhancing its predicting accuracy. The remaining 30\% served as an independent testing set to determine the efficacy of our proposed model  on unseen data.

\begin{figure}
    \centering
    \includegraphics[width=1\linewidth]{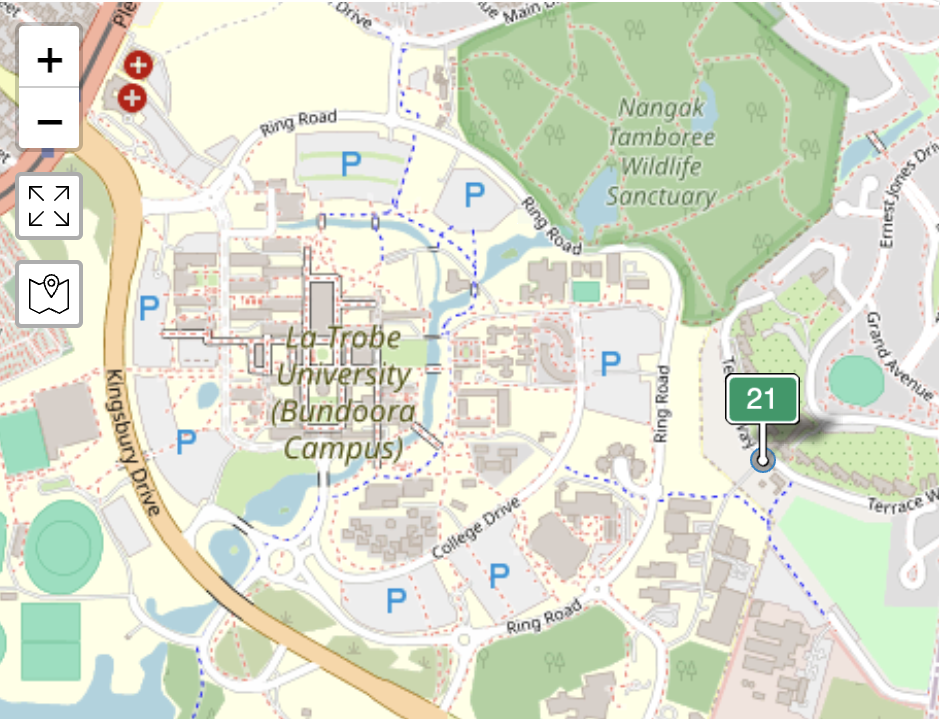}
    \caption{Location where dataset was sourced from (La Trobe University) along with nearest AQI station (Macleod, Victoria)}
    \label{fig:enter-label}
\end{figure}
\begin{figure}
    \centering
    \includegraphics[width=1\linewidth]{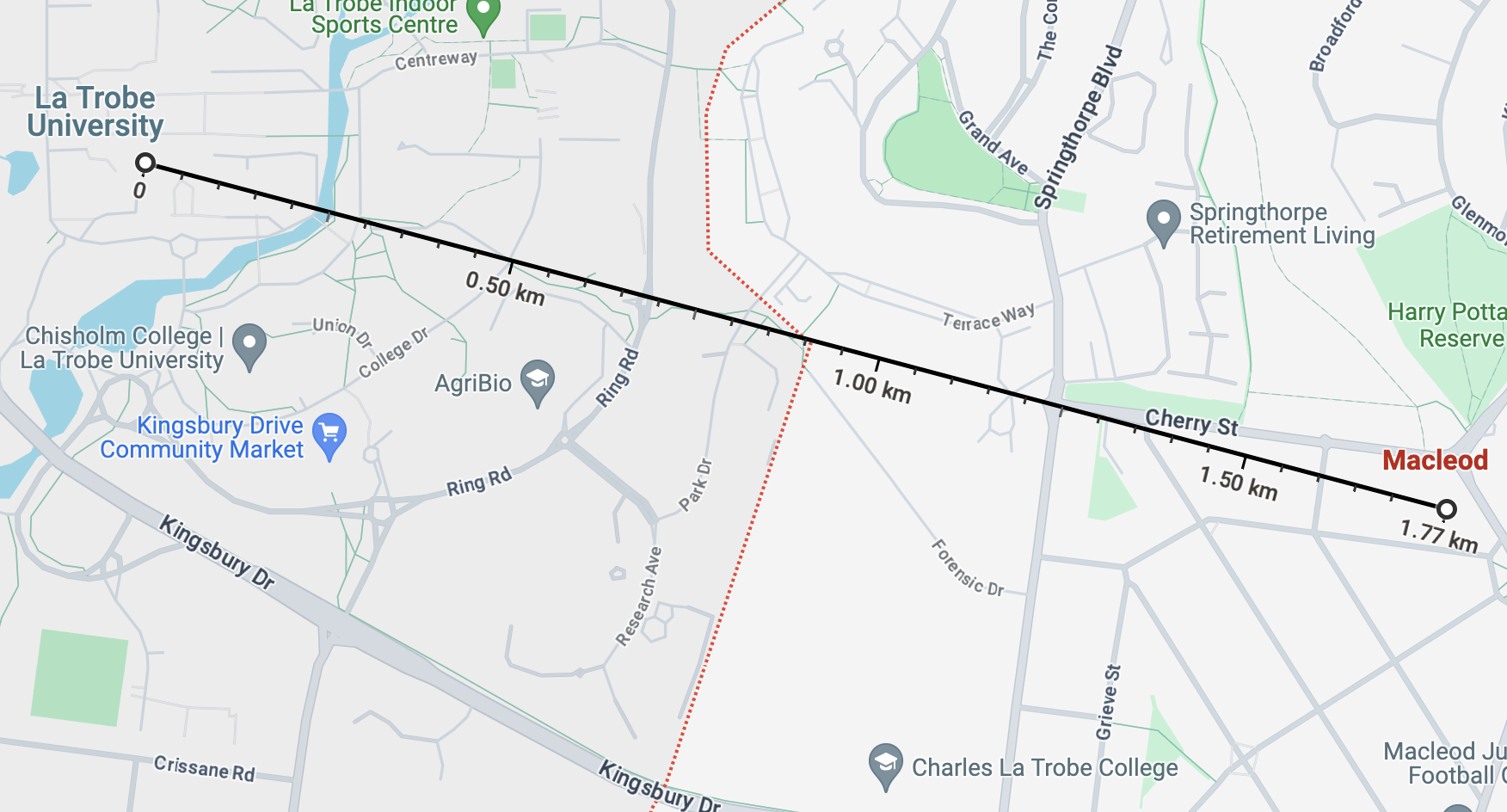}
    \caption{Distance in kilometers between La Trobe University and Macleod AQI station (1.77km)}
    \label{fig:enter-label}
\end{figure}

\begingroup
\small 

\endgroup

\subsection{Time-Series Approach}

Initially, a regression-based approach was utilized to predict the solar power generation based on the factors present. However, this did not provide adequate information regarding the relationship between these factors and solar power generation.\\
This prompted us to try out a time series-based approach as we also had chronological data. This is because a time series model is designed to capture patterns in sequential and discrete data points over a set period in time. We made use of solar generation at a particular moment in time to predict the generation at some time in the future as this type of approach would help us identify trends and seasonality to make more accurate predictions. We decided to shape our prediction by creating a column that would predict solar generation values 24, 48 and 72 hours out and compared the effectiveness of these models.\\
The Machine Learning models used for generating and comparing solar power generation for a Time Series approach were:

\begin{itemize}
\item \textbf{Linear Regression}\\It is a statistical model used to establish a linear relationship between a dependent variable and one or more independent variables. It allows for the prediction of coefficients that decide the strength of the established linear relationship. It is used to predict the line of best fit which minimizes the distance between actual values and predicted values.
\\ 
\begin{equation} 
Y = \beta_0 + \beta_1X_1 + \beta_2X_2 + \ldots + \beta_pX_p + \varepsilon
\end{equation}




\item \textbf{Gradient Boosting Regression}\\It is an ensemble model which combines multiple weak prediction models like decision trees,
and takes the strongest combination of predictions to build a strong predictive model. Gradient Boosting works by repeatedly training weak models to fit the negative gradient of the ongoing prediction’s loss model, which helps improve its accuracy.
\\
\begin{equation}
\hat{y}_i = \sum_{k=1}^{K} f_k(x_i) = f_0(x_i) + \sum_{k=1}^{K} \gamma_k h_k(x_i)
\end{equation}

\item \textbf{Random Forest Regression}\\It is a supervised learning algorithm used for regression-based tasks. It combines concepts of decision trees and ensemble learning to make precise predictions. It also reduces overfitting due to an element of randomness which restricts individual trees from memorizing the training set data provided to it. The decision of the algorithm is made by aggregating the decisions of individual trees using majority voting.
\\
\begin{equation}
\hat{y}_i = \frac{1}{N} \sum_{j=1}^{N} f_j(x_i)
\end{equation}

\item \textbf{XGBoost Regression}\\Extreme Gradient Boosting is an advanced gradient boosting algorithm that is widely used due to the high accuracy of predictions. It achieves this by iteratively adding weak models to the ensemble and ensuring they fit according to the current prediction. It also uses features like column and row subsampling to further improve its performance.
\vspace{-3pt}
\begin{equation}
\hat{y}_i = \sum_{k=1}^{K} f_k(x_i) = f_0(x_i) + \sum_{k=1}^{K} h_k(x_i)
\end{equation}

\vspace{4.5pt}

    \item \textbf{Random Forest Regression + XGBoost Regression}\\
    Building upon the positive results displayed by Lokesh et al [20], we decided upon an ensemble model consisting of a Random Forest Regressor and a XGBoost Regressor, considering that their strengths are complementary in the sense that both are ensemble learners and use boosting and bagging respectively. RFR can handle non-linear relationships between data points and XGBoost can capture subtle patterns and temporal dependencies.

    \vspace{4.5pt}
    \item \textbf{ConvLSTM2D}\\
In this subsection, we delve into the specifics of the ConvLSTM2D model, a hybrid architecture that merges CNNs and LSTM networks. This unique architecture is designed specifically for analyzing spatio-temporal data, where understanding both spatial relationships and temporal dynamics is crucial for accurate predictions.

\begin{figure}[H]
\centering
\includegraphics[width=1\linewidth]{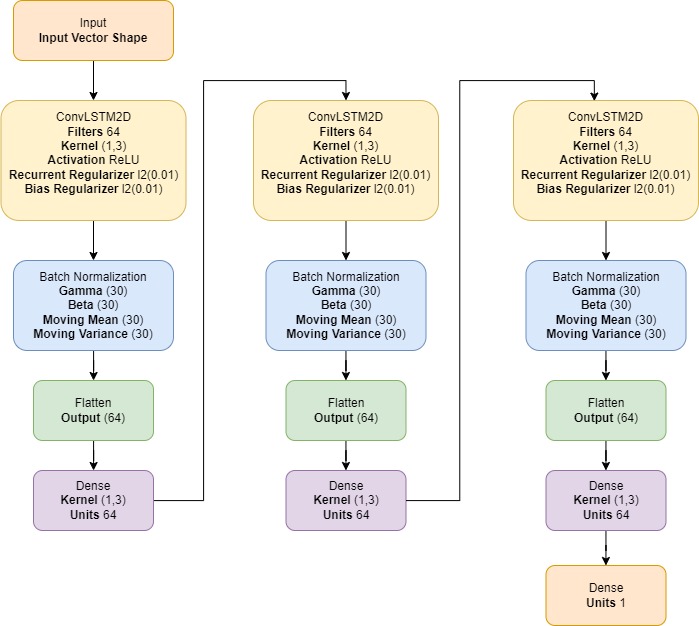}
\caption{This figure shows the ConvLSTM2D model architecture, which combines convolutional and LSTM layers to process spatio-temporal data.}
\label{fig:convlstm2d_architecture}
\end{figure}
The architecture of the ConvLSTM2D model, illustrated in Figure 3, consists of layers that facilitate the processing of spatio-temporal data. At its core are ConvLSTM units, which extend the traditional LSTM cells by incorporating convolutional operations within the recurrent structure. This innovative design enables the model to effectively capture both spatial and temporal dependencies within the data. The utilization of ConvLSTM units makes the model adept at handling sequences of spatio-temporal observations, a characteristic essential for tasks such as video analysis, weather forecasting, and motion prediction.

The training procedure of the ConvLSTM2D model follows a systematic approach tailored to exploit its full potential. Initially, standard data preprocessing techniques, similar to those employed for other machine learning models, are applied. However, due to the unique architecture of ConvLSTM2D, additional reshaping of the data is necessary to conform to its input shape requirements. The input data is reshaped into a 5-dimensional tensor, accommodating batches of sequences, each comprising 2D matrices over time. This reshaping operation enables the model to interpret the input data as a spatio-temporal sequence, facilitating effective learning of complex patterns. Furthermore, to ensure robust training and prevent issues stemming from skewed distributions, a power transformation is applied to scale input features appropriately. This normalization step fosters homogeneous feature scales, thereby preventing any individual feature from disproportionately influencing the learning process.

During the training phase, the model is optimized using the Adam optimization algorithm, with a predefined learning rate. The optimization process involves iteratively adjusting the model's internal parameters to minimize the mean squared error loss between predicted and actual values. This iterative optimization enables the model to discern intricate patterns within the data and refine its predictive capabilities accordingly.

The ConvLSTM2D model was chosen for its inherent capability to capture spatio-temporal dependencies effectively, a critical requirement in our analysis. Through rigorous experimentation and evaluation, it consistently outperformed alternative models, exhibiting superior performance in terms of predictive accuracy and generalization capabilities. Its hybrid architecture, which integrates both convolutional and recurrent operations, enables it to discern complex patterns within spatio-temporal data, making it particularly well-suited for the tasks at hand.

\end{itemize}

\subsection{Zero-Inflated Models}

An analysis of the solar power data showed that the distribution was such that an unusually high number of values amounted to zero, this was likely due to negligible solar generation due to the intermittent nature of sunlight received by the solar panels. These zeros overshadowed the other values in the distribution by a fair amount. 
\\
The histogram shown in Figure 2 highlights this zero-inflated data of our initial dataset. It also provides insights into the range of values of solar energy generation and their respective frequencies.
\begin{figure}[htbp]
    \centering
    \includegraphics[width=1\linewidth]{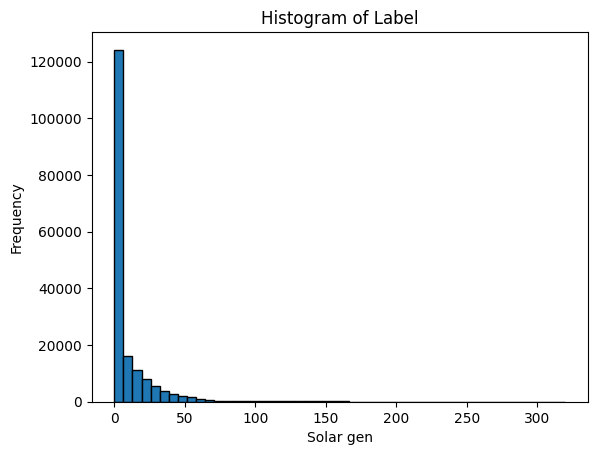}
    \caption{Histogram of the initial solar power distribution before application of transformations, highlighting the zero-inflated nature of our dataset.}
    \label{fig:1}
\end{figure}
\\
\\
Thus, on noticing this skewness of the solar energy generation data, we decided to switch to a zero-inflated model which essentially differentiates between structural zeros(due to nighttime and genuine absences of solar generation) and zero inflation(additional factors that may influence the probability of observing a zero, such as cloudy days or equipment failures).
According to a study, previous research indicates that if excessive zero is not accounted for, an unreasonable fit for both the zeros and nonzero counts will occur (Perumean-Chaney et al. 2013). ZI (Lambert 1992) and hurdle models(Mullahy 1986; Heilbron 1994) have been developed to model zero-inflation when the regular count models such as Poisson or negative binomial are unrealistic'[17]. Thus it became a necessity to apply such a zero-inflated model to our data that would best represent the state of the current distribution and convert it into a distribution that might be more acceptable for our regression models to make predictions with. After considering various models and comparing them with our data to determine the best fit, we decided upon the Tweedie distribution.

\textbf{Tweedie Distribution}

The Tweedie distribution(Figure 3) is characterized by the following components:
\begin{itemize}
    \item Random Variable: Let $y$ be the random variable representing the observed value.
    \item Mean Parameter: $\mu$ is the mean parameter of the distribution, indicating the average value of $y$.
    \item Power Parameter: $p$ is the power parameter of the distribution, controlling the variance structure. It determines the shape of the distribution and can take any positive value, excluding 1.
    \item Normalizing Constant: $A(\mu, p)$ is the normalizing constant that ensures the Probability Density Function (PDF) integrates to 1 over the support of $y$. It accounts for the specific value of $\mu$ and $p$ and is essential for properly defining the distribution.
\end{itemize}
The probability density function of the Tweedie distribution is thus given by:

\begin{align*}
f(y; \mu, p) = \frac{y^{p-1} \exp\left(\frac{\mu(1-p)}{1-p}\right)}{A(\mu, p)} \exp\left(-\frac{y^{p}}{A(\mu, p)}\right) 
\end{align*}
\begin{figure}[htbp]
    \centering
    \includegraphics[width=1\linewidth]{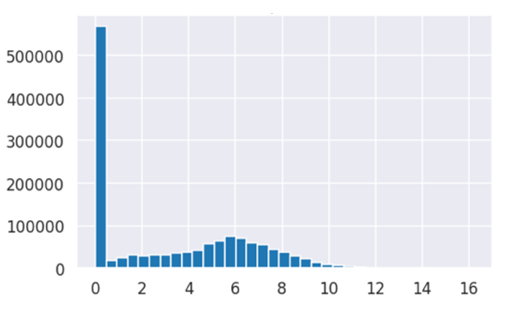}
    \caption{A plot demonstrating a standard Tweedie distribution.}
    \label{fig:3}
\end{figure}

The Tweedie distribution encompasses various shapes, ranging from heavy-tailed with excess zeros (for $p < 1$) to symmetric and Gaussian-like (for $1 < p < 2$ and $p > 2$). It includes special cases such as the Poisson distribution (when $p = 1$) and the gamma distribution (when $p = 2$). The choice of $p$ determines the specific characteristics of the Tweedie distribution, including its skewness, tail behavior, and overall shape. This was suitable for our zero-inflated distribution, by using a modified version of the above called a Zero Inflated Tweedie (ZIT) Model which accounts for the excess zeros using a separate inflation component.
\\

\textbf{Zero-Inflated Model with Tweedie Distribution}

The zero-inflated model with the Tweedie distribution can be represented using the following formula:

\begin{align*}
Y &= \begin{cases}
0, & \text{with probability } \pi \\
Z, & \text{with probability } 1 - \pi
\end{cases} \\
\log\left(\frac{\mu}{\phi}\right) &= \beta_0 + \beta_1 X_1 + \beta_2 X_2 + \ldots + \beta_p X_p \\
\log(\phi) &= \gamma_0 + \gamma_1 X_1 + \gamma_2 X_2 + \ldots + \gamma_q X_q \\
\log\left(\frac{\pi}{1 - \pi}\right) &= \theta_0 + \theta_1 X_1 + \theta_2 X_2 + \ldots + \theta_r X_r \\
\end{align*}

where:
\\
- \(Y\) represents the response variable (count variable with excess zeros).
\\
- \(Z\) represents the positive count variable.
\\
- \(\pi\) represents the probability of excess zeros.
\\
- \(\mu\) represents the mean parameter.
\\
- \(\phi\) represents the dispersion parameter.
\\
- \(X_1, X_2, \ldots, X_p\) represent the predictor variables for the mean equation.
\\
- \(X_1, X_2, \ldots, X_q\) represent the predictor variables for the dispersion equation.
\\
- \(X_1, X_2, \ldots, X_r\) represent the predictor variables for the zero-inflation equation.
\\
- \(\beta_0, \beta_1, \beta_2, \ldots, \beta_p\) represent the coefficients for the mean equation.
\\
- \(\gamma_0, \gamma_1, \gamma_2, \ldots, \gamma_q\) represent the coefficients for the dispersion equation.
\\
- \(\theta_0, \theta_1, \theta_2, \ldots, \theta_r\) represent the coefficients for the zero-inflation equation.
\\

Now in order to accurately and efficiently apply a customized zero-inflated model to our data and test out various regression models in Python, we utilized H2O, a Java-based software for data modeling and general computing.
\\
H2O is an abstracted distributed processing engine that allows for simple horizontal scaling in order to provide solutions faster and more efficiently. When it comes to model application, it consists of numerous estimator functions like H2OGradientBoostingEstimator and H2ODeepLearningEstimator served using REST API abstraction, each of which consists of a plethora of hyperparameters for deep customization. For the scope of our project, we have chosen  H2OGradientBoostingEstimator, H2OXGBoostEstimator, H2ODeepLearningEstimator, and H2ORandomForestEstimator. 
The value of the parameters \textit{distribution}  and 
\( \textit{tweedie\_power} \) were set to \textit{‘tweedie’} and \textit{‘1.5’} respectively for the models and other parameters were optimized using GridSearchCV. For the Deep Learning Zero-inflated MOdel (ZIM), 4 hidden layers were chosen of neuron count 100,100,50 and 50 respectively. 
Transforming our data using a zero-inflated model resulted in a marked improvement in our solar power prediction with a reduced MAE and RMSE.

\subsection{Power-Transform}
It is a data transformation and scaling technique which is another way to tackle the skewness of the solar generation data; we found that using PowerTransformer was a significantly better fit for our dataset than the zero-inflated model. This scaling method is applied feature-wise to make the data more Gaussian or Gaussian-like which is inherently assumed by regression-based prediction models. It is used when dealing with non-constant variance. There are 2 different methods of  performing the power transform, namely the Box-Cox transform and the Yeo-Johnson transform.
\\
The Yeo-Johnson power transform is given by the formula:
\[
x_{\lambda} = 
\begin{cases}
    \left((x + 1)^{\lambda} - 1\right)/\lambda, & \text{if } \lambda \neq 0, x \geq 0 \\
    \ln(x + 1), & \text{if } \lambda = 0, x \geq 0 \\
    -\left((|x| + 1)^{2 - \lambda} - 1\right)/(2 - \lambda), & \text{if } \lambda \neq 2, x < 0 \\
    -\ln(|x| + 1), & \text{if } \lambda = 2, x < 0 \\
\end{cases}
\]

Here, \(x\) represents the original variable, and \(\lambda\) is a parameter that determines the type of power transform applied. The transformed variable \(x_{\lambda}\) is the result of applying the Yeo-Johnson power transform.
The aforementioned Yeo-Johnson power transform was thus applied to the data, and the resulting distribution was notably found to resemble a Tweedie distribution.
\begin{figure}[htbp]
    \centering
    \includegraphics[width=1\linewidth]{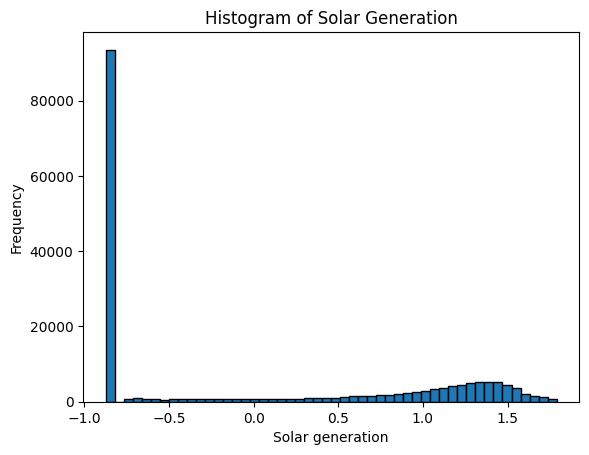}
    \caption{Histogram of solar power distribution after applying power transform which is analogous to the tweedie distribution in figure 5.}
    \label{fig:1}
\end{figure}

The solar energy generation data points are now normalized to make the distribution more Gaussian as shown in Figure 4.

\begin{figure*}[btp]
    \centering
    \includegraphics[width=0.8\linewidth]{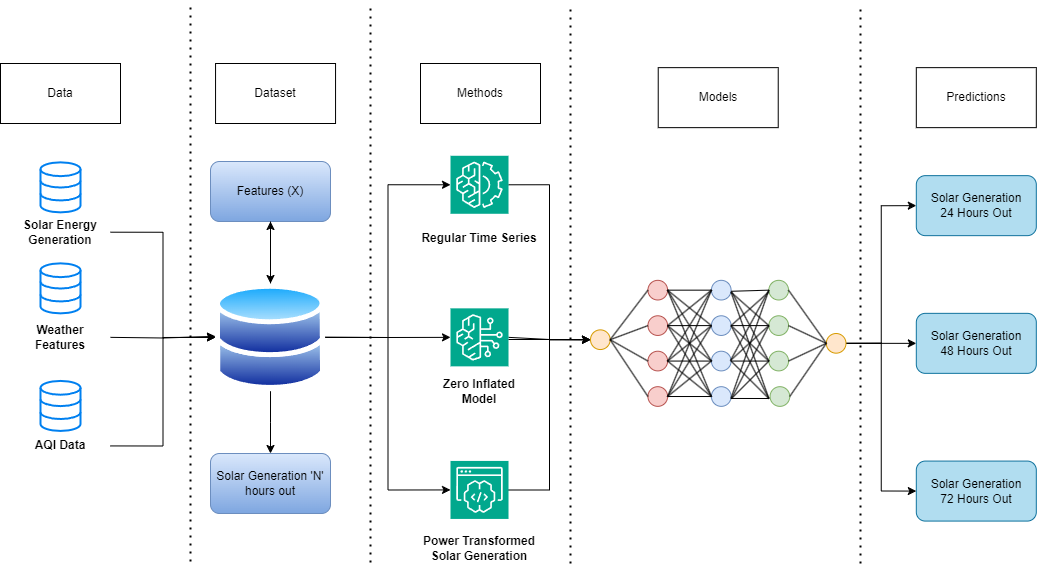}
     \caption{Complete pipeline used for data collection and prediction of solar data.}
    \label{fig:1}
\end{figure*}

\section{Results}
For the prediction of solar energy generation using multiple methodologies, we have found that the Power Transformed data led to the most accurate prediction in comparison to Regular Time Series and Zero-Inflated models. Power Transformation of data is a particular method that stands out in comparison to the rest. This is due to solar energy generation being dependent on various factors like temperature, seasonality, time of day, and air quality of the region. These data points have non-linear relationships with each other. ConvLSTM2D models outperform normal regression models as it combines convolutional operations and LSTM memory cells, allowing for the modeling of both spatial and temporal dependencies in data. This makes ConvLSTM2D well-suited for tasks where both spatial and sequential information are important, such as solar power generation forecasting.

For evaluating the performance of the models on the gathered data, various statistical metrics were considered, of which ultimately R\(^2\), MAE, and RMSE were used. Our key metric was R\(^2\), as the coefficient of determination R\(^2\) is generally a better indicator of regression model performance when compared to other metrics[18]. These metrics were tabulated and compared across the various models and for three different time slots: 24 hours, 48 hours and 72 hours. There are three such tables corresponding to the principal methodologies utilised: Regular time-series, Zero Inflated Model, and Power Transform.

\begin{table}[!ht]
    \centering
    \caption {Evaluation metrics applied for regular time series approach}
    \begin{tabular}{|c|c|c|c|c|}
    \hline
        Models & Hours Out & R\(^2\)
 Score & MAE & RMSE \\ \hline
        \cline{2-5}
        ~ & 24 & 0.6344 & 6.45 & 14.91 \\ 
        \cline{2-5}
        Linear Regression & 48 & 0.5884 & 7.08 & 15.99 \\ 
        \cline{2-5}
        ~ & 72 & 0.5677 & 7.57 & 16.43 \\ \hline

        ~ & 24 & 0.7142 & 5.15 & 13.18 \\ 
        \cline{2-5}
        GradientBoosting Regression & 48 & 0.6793 & 5.62 & 14.12 \\ 
       \cline{2-5}
        ~ & 72 & 0.6555 & 6.13 & 14.66 \\ \hline
        ~ & 24 & 0.7431 & 5.00 & 12.49 \\ 
        \cline{2-5}
        XGBoost Regression & 48 & 0.7095 & 5.47 & 13.44 \\ 
        \cline{2-5}
        ~ & 72 & 0.6865 & 5.92 & 13.99 \\ \hline
        ~ & 24 & 0.8005 & 3.42 & 11.02 \\ 
        \cline{2-5}
        RandomForest Regression & 48 & 0.7987 & 3.52 & 11.19 \\ 
        \cline{2-5}
        ~ & 72 & 0.7996 & 3.48 & 11.18 \\ \hline
        ~ & 24 & 0.8244 & 3.49 & 10.33 \\ 
        \cline{2-5}
        RandomForest + XGBoost & 48 & 0.8145 & 3.71 & 10.74 \\ 
        \cline{2-5}
        ~ & 72 & 0.8178 & 3.75 & 10.66  \\ \hline
    \end{tabular}
\end{table}

\begin{table}[!ht]
    \centering
    \caption {Evaluation metrics applied for Zero Inflated Model time series approach}
    \begin{tabular}{|c|c|c|c|c|}
    \hline
        Models & Hours Out & R\(^2\)
 Score & MAE & RMSE \\ \hline
        \cline{2-5}
        ~ & 24 & 0.7979 & 3.99 & 11.07 \\ 
        \cline{2-5}
        GradientBoosting Regression & 48 & 0.7678 & 4.37 & 11.90 \\ 
       \cline{2-5}
        ~ & 72 & 0.7506 & 4.55 & 12.29 \\ \hline
        ~ & 24 & 0.7790 & 4.17 & 11.58 \\ 
        \cline{2-5}
        XGBoost Regression & 48 & 0.7501 & 4.54 & 12.35 \\ 
        \cline{2-5}
        ~ & 72 & 0.7406 & 4.76 & 12.61 \\ \hline
        ~ & 24 & 0.8086 & 3.51 & 10.65 \\ 
        \cline{2-5}
        RandomForest Regression & 48 & 0.7987 & 3.63 & 10.74 \\ 
        \cline{2-5}
        ~ & 72 & 0.8165 & 3.62 & 10.61 \\ \hline
        ~ & 24 & 0.6761 & 4.99 & 14.00 \\ 
        \cline{2-5}
        Deep Learning & 48 & 0.6287 & 5.55 & 16.26 \\ 
        \cline{2-5}
        ~ & 72 & 0.6285 & 5.33 & 16.18  \\ \hline
    \end{tabular}
\end{table}

The range of target variables was between 0 and 320 for regular time series and Zero-Inflated models. For the Zero Inflated Model, we performed five-fold cross-validation, and there was no significant variance in the validation accuracy from the training accuracy.

\begin{table}[!ht]
    \centering
    \caption {Evaluation metrics applied for Power Transform time series approach}
    \begin{tabular}{|c|c|c|c|c|}
    \hline
        Models & Hours Out & R\(^2\)
 Score & MAE & RMSE \\ \hline
        \cline{2-5}
        ~ & 24 & 0.8357 & 0.21 & 0.41 \\ 
        \cline{2-5}
        Linear Regression & 48 & 0.7875 & 0.25 & 0.46 \\ 
        \cline{2-5}
        ~ & 72 & 0.7380 & 0.30 & 0.51 \\ \hline
        ~ & 24 & 0.8754 & 0.18 & 0.35 \\ 
        \cline{2-5}
        GradientBoosting Regression & 48 & 0.8345 & 0.21 & 0.41 \\ 
       \cline{2-5}
        ~ & 72 & 0.7952 & 0.25 & 0.45 \\ \hline
        ~ & 24 & 0.8926 & 0.17 & 0.33 \\ 
        \cline{2-5}
        XGBoost Regression & 48 & 0.8581 & 0.20 & 0.38 \\ 
        \cline{2-5}
        ~ & 72 & 0.8289 & 0.24 & 0.41 \\ \hline
        ~ & 24 & 0.9574  & 0.09 & 0.21 \\ 
        \cline{2-5}
        RandomForest Regression & 48 & 0.9580 & 0.09 & 0.21 \\ 
        \cline{2-5}
        ~ & 72 & 0.9584 & 0.08 & 0.20 \\ \hline
        ~ & 24 & 0.9595 & 0.09 & 0.20 \\ 
        \cline{2-5}
        RandomForest + XGBoost & 48 & 0.9561 & 0.21 & 0.10 \\ 
        \cline{2-5}
        ~ & 72 & 0.9562 & 0.11 & 0.21  \\ \hline
    \end{tabular}
\end{table}
The range of target variables was between -0.9 and 1.8 after Power Transform was applied.
\\
\begin{table}[!ht]
    \centering
    \caption {Evaluation metrics applied for LSTM time series approach}
    \begin{tabular}{|c|c|c|c|c|}
    \hline
        Model & Hours Out & R\(^2\)
 Score & MAE & RMSE \\ \hline
        \cline{2-5}
        ~ & 24 & 0.9691 & 0.18 & 0.10 \\ 
        \cline{2-5}
        ConvLSTM2D & 48 & 0.9637 & 0.18 & 0.08 \\ 
       \cline{2-5}
        ~ & 72 & 0.9608 & 0.20 & 0.09 \\ \hline
    \end{tabular}
\end{table}

After comparing all three methodologies, regression models, and variation in prediction time, the best combination of factors respectively has been tabulated in Table V.
\begin{figure}[h]
  \centering
  \includegraphics[width=0.75\linewidth]{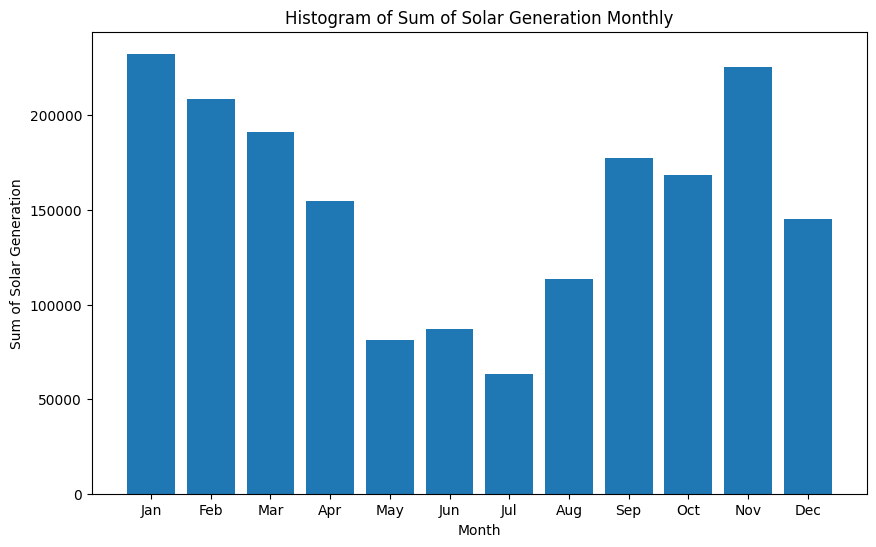}
  \caption{Monthly solar generation which provides insights into the seasonality of the data}
  \label{fig:solar-generation}
\end{figure}

\begin{table}[!ht]
    \centering
    \caption {Optimum metric for Power Transform time series approach}
    \begin{tabular}{|c|c|c|c|c|}
    \hline
        Model & Hours Out & R\(^2\)
 Score & MAE & RMSE \\ \hline
        \cline{2-5}
        RandomForest + XGBoost & 24 & 0.9595 & 0.09 & 0.20 \\ \hline
        \cline{2-5}
        ConvLSTM2D & 24 & 0.9691 & 0.18 & 0.10 \\ \hline
    \end{tabular}
\end{table}

\begin{figure}
    \centering
    \includegraphics[width=0.8\linewidth, height=5cm]{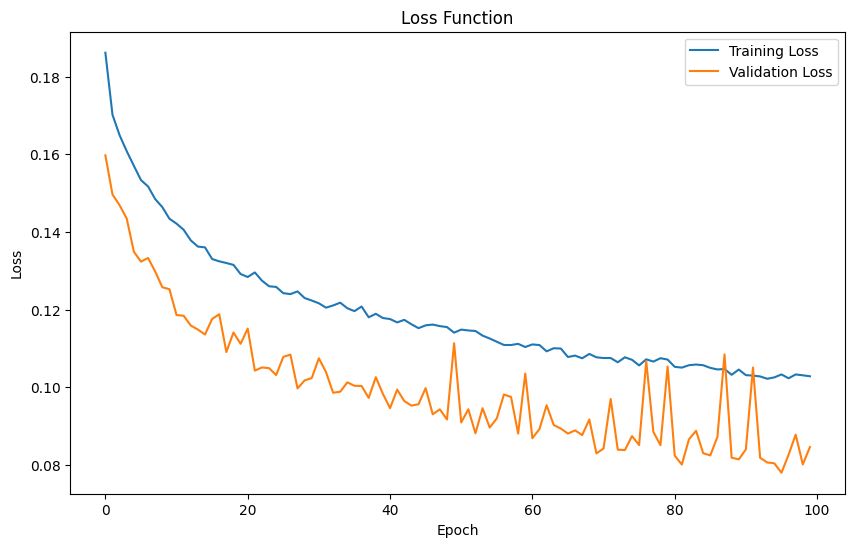}
    
    \caption{Loss vs Epoch for ConvLSTM2D}
    \label{fig:1}
\end{figure}

The solar power generation data when plotted monthly follows a specific pattern that can be attributed to the seasonal cycle of the Australian landmass, where the dataset was sourced from. The generation is noted to be maximum from November to February which coincides with the summer months in Australia and reaches its minimum during the months of May to August which are the winter months. This can be explained due to the sun being directly overhead during summer, leading to longer days and more exposure to solar radiation. This phenomenon reverses during the winter months as is shown by the histogram(Fig. 6).

\begin{figure}[!ht]
  \centering
  \includegraphics[width=1\linewidth]{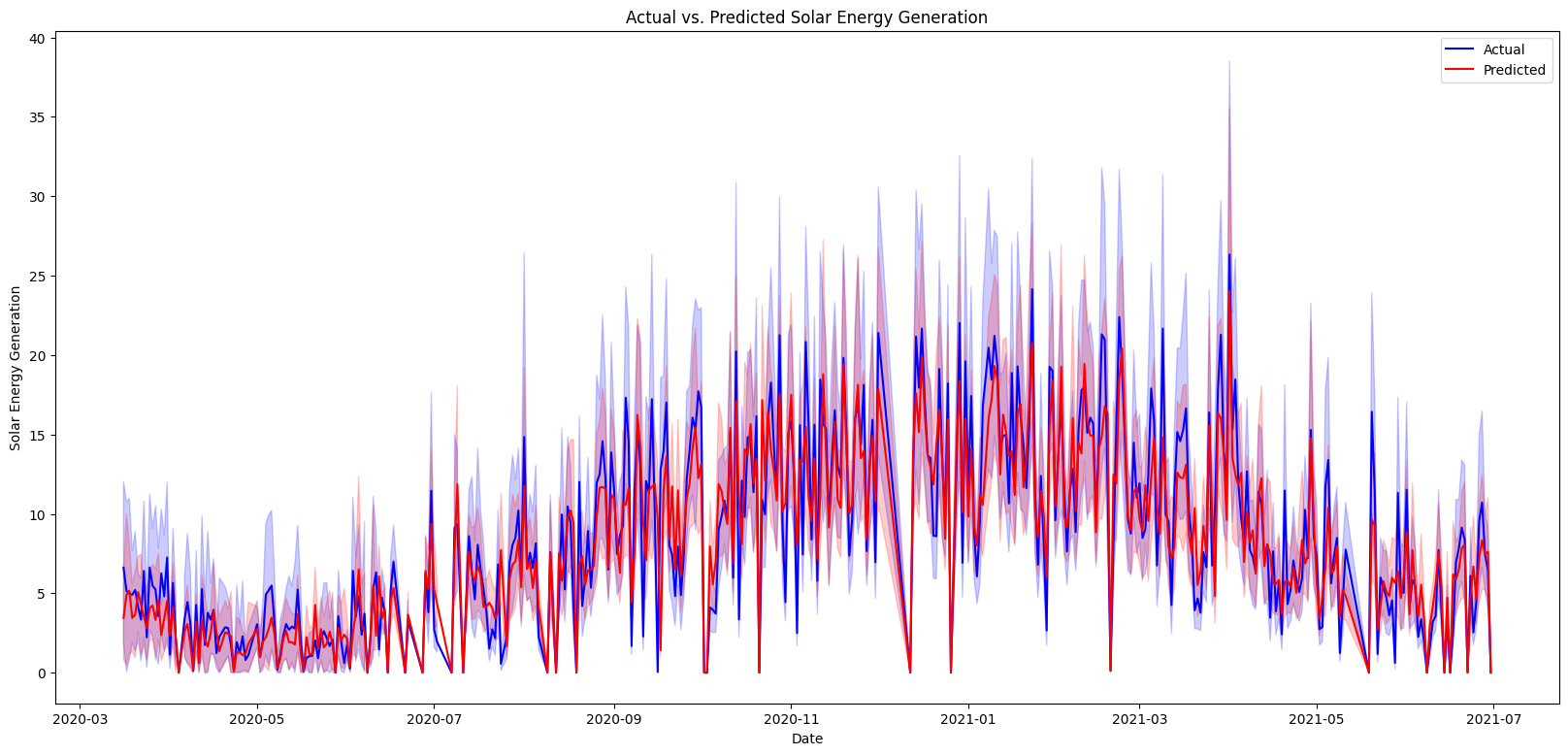}
  \caption{Zero Inflated 24H out prediction using Gradient Boost Regressor}
  \label{fig:zero-inflated}
\end{figure}

\begin{figure}[!ht]
    \centering
    \includegraphics[width=1\linewidth]{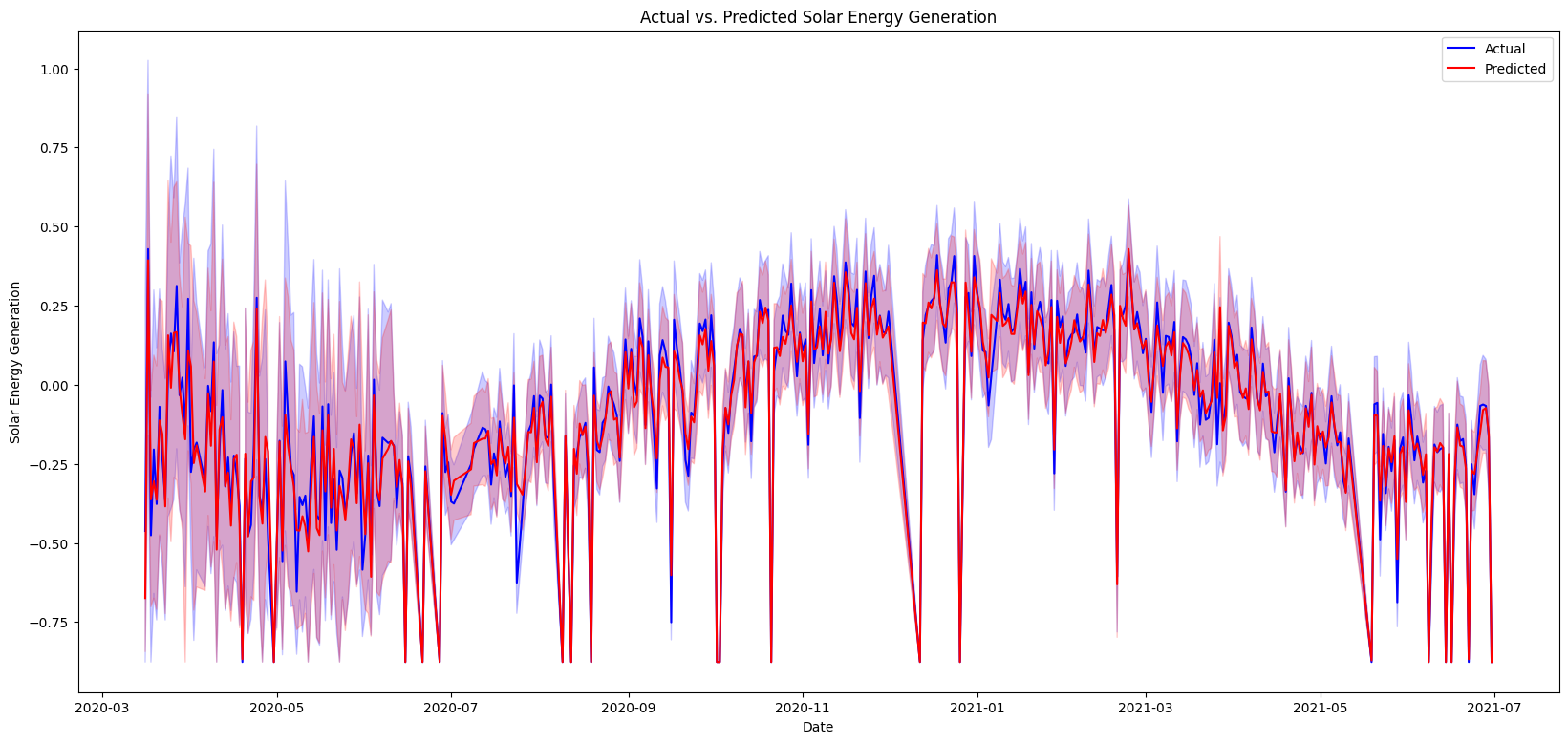}
    \caption{Power Transformed 24H out prediction using RandomForest Regressor}
    \label{fig:enter-label}
\end{figure}
Figure 7 shows the Loss vs Epoch plot for the ConvLSTM2D model displaying the training progress.
\\
\\
Our time-series based solar power prediction models also capture this phenomenon as seen in the graphs plotted above (Fig. 8) and (Fig. 9). There is a drop in solar production during the winter months of 2020 and the peak production is reached in the summer of 2021.

\section{Conclusion}
This study investigates the use of various machine learning algorithms to effectively determine the future solar generation in a region by utilizing a time series approach.
The chief models employed were Linear Regression, Lasso, Ridge, ElasticNet, ensemble models like RandomForest and XGBoost, and deep learning models like ConvLSTM2D. Later, on conferring, we decided to switch to a time- series based approach due to the seasonal fluctuation in the solar data. However, on noticing the skewness of the solar energy generation data which had a high number of zeros, we decided to switch to a zero-inflated model which helps ascertain the difference between the true zero data points and the inflated zeros. This approach immediately yielded a higher accuracy of solar prediction with a lower mean standard error. Another way to tackle the skewness of the solar generation data was to try different scaling techniques; we found that using PowerTransformer was a significantly better fit for our dataset than the zero-inflated model. This scaling method is applied feature-wise to make the data more Gaussian or Gaussian-like which is inherently assumed by regression-based prediction models.
In conclusion, this study investigates the use of machine learning algorithms incorporating AQI and climate factors to provide more accurate solar generation forecasts. Considering seasonal variations in the solar data, the time series-based method was modified. In addition, a zero-inflated model and scaling techniques were used to address the skewness of the solar generation data. The findings provide valuable insights for solar stakeholders, contributing to the adoption and use of sustainable energy sources.

\section{Future Scope}
Solar energy generation forecasting is a dynamic field that will always develop and demand exploration. There is a lot of scope in the future to improve the accuracy of solar forecasting. To make this forecasting scalable and generalized for all geographical regions, the dataset can be collected from different regions and time periods since our research primarily utilizes data from a specific region and time zone. Additionally, data fusion and feature engineering can be done to enhance the power of forecasting. Data fusion is the practice of collecting data from multiple sources like satellite imagery and weather stations which can be combined together for better results. There was a limitation on the dataset which we faced during implementation which was the fact that the AQI data we used wasn't from a data station at the exact geographical location of the data collection but rather another data station away from the solar site. Thus in order to increase accuracy one can take AQI data from a data station at maximum proximity to the data collection center in order to eliminate many geographical errors. Another possibility can be to collect our own data to create a localized dataset which would likely be more accurate since we can customize it and remove any bottlenecks faced earlier by having utilised an external dataset. Moreover, feature engineering can be employed to extract more meaningful features from the data available, hence increasing the model performance. Features like cloud cover, dust, and further manners of seasonality can also be taken into consideration for future research.

\end{document}